# Líneas de investigación en minería de datos en aplicaciones en ciencia e ingeniería: Estado del arte y perspectivas.

José A. García Gutiérrez,
*Universidad Nacional de Educación a Distancia,*
*Calle Juan del Rosal,*
*29040 Madrid – España*
*Jgarcia5053 @*
*alumno.uned.es*

The continuous increase in the availability of data of any kind, coupled with the development of networks of high-speed communications, the popularization of cloud computing and the growth of data centers and the emergence of high-performance computing does essential the task to develop techniques that allow more efficient data processing and analyzing of large volumes datasets and extraction of valuable information. In the following pages we will discuss about development of this field in recent decades, and its potential and applicability present in the various branches of scientific research. Also, we try to review briefly the different families of algorithms that are included in data mining research area, its scalability with increasing dimensionality of the input data and how they can be addressed and what behavior different methods in a scenario in which the information is distributed or decentralized processed so as to optimize performance in heterogeneous environments.

El aumento continuo de la disponibilidad de datos de toda naturaleza, unido al desarrollo de redes de comunicaciones de alta velocidad, la intercomunicación de los centros de datos y la aparición de la computación de alto desempeño, hace imprescindible el desarrollo de técnicas de minería de datos que permitan procesar y analizar grandes volúmenes de datos y extraer de ellos información de valor. En las siguientes páginas hablaremos sobre la evolución que ha tenido este campo en las últimas décadas, así como de su potencialidad y aplicabilidad presente en las diferentes ramas de la investigación científica. Así mismo, trataremos de repasar de forma breve las diferentes familias de algoritmos que se engloban en el campo de la minería de datos, su escalabilidad cuando aumenta la dimensionalidad de los datos de entrada y de cómo se puede abordar y cuál es el comportamiento de los diferentes métodos en un escenario en el que la información se encuentra distribuida o se procesa de manera descentralizada o paralela de forma que se pueda optimizar el rendimiento en entornos heterogéneos.

## 1. INTRODUCTION

En las dos últimas décadas, y de forma paralela al desarrollo de los sistemas de información, se ha puesto en valor el conocimiento como recurso estratégico y las bases de datos con información transaccional han pasado a convertirse en valiosos repositorios para las técnicas minería de datos que pasan a ser herramientas fundamentales en cualquier ámbito científico o empresarial ya que permiten obtener información clave permitiendo analizar, extraer patrones y categorizar los datos de forma que se han hecho imprescindibles en cualquier proyecto de investigación.

Entre los usos que hoy día se dan a las técnicas de minería de datos encontramos: la extracción de reglas o interrelaciones entre los diferentes atributos que conforman un conjunto de datos, la predicción de valores o series, la inferencia bayesiana, las tareas de asistencia a la toma de decisiones y los procesos de clasificación, agrupación o caracterización de un conjunto de datos, entre otros. Hasta el desarrollo de las técnicas de aprendizaje automático, la manera de abordar un gran volumen de datos pasaba por realizar un acercamiento estadístico basado en métodos procedentes en su mayoría de la estadística o las matemáticas.

No es hasta finales de los 80 con el desarrollo de los computadores personales y la llegada de estos a todos los ámbitos de la sociedad (lo que provocó la digitalización de una parte importante de los procesos de negocio) que se empieza a hablar del desarrollo de bases de datos para el análisis en línea o OLAP (del inglés *Online Analytical Processing*) y de las primeras técnicas de Descubrimiento de Conocimiento en Bases de Datos (o KDD por sus siglas en inglés acrónimo de *Knowledge Discovery in Databases*) y Minería de Datos (del inglés *Data Mining*, en adelante MD).

Ambos términos, incluso hoy en día, son utilizados frecuentemente de forma incorrecta para referirse a la extracción de información de grandes volúmenes de datos (sobre todo desde el enfoque de la inteligencia empresarial). Sin embargo, y aunque ambos términos no son equivalentes (la Minería de Datos es un paso dentro del proceso de KDD, el cual a su vez es una forma de Aprendizaje Automático), se podría decir que el proceso de Minería de Datos es el paso más importante dentro de este proceso. Tal y como exponen varios autores en la literatura [Berry et al. 2004, Hernández et al. 2004, Mitra et al. 2003, MacLennan et al. 2008, Riquelme et al., 2006] podemos definir las tareas de descubrimiento de conocimiento en BBDD como:

---

(1) Se puede consultar el dosier de prensa del lanzamiento de la misión Gaia en el website de la Agencia Espacial Europea en url http://www.esa.int/esl/ESA_in_your_country/Spain/Dosier_de_Prensa_-_Mision_Gaia_-

(2) el término original en inglés (americano) es "billion", este fue traducido al español "mil millones", debido a que en español (según la Real Academia Española de la Lengua), "1 billón" equivale a "un millón de millones".

MAS NOTAS AL FINAL DEL DOCUMENTO.

"El Descubrimiento de Conocimiento en bases de datos (KDD) puede definirse como el proceso de identificar patrones significativos en los datos que sean válidos, novedosos, potencialmente útiles y comprensibles para un usuario".

Aunque existen varias definiciones en la bibliografía sobre Minería de Datos focalizadas en diferentes áreas de estudio (obtención de patrones de comportamiento, inteligencia empresarial, análisis de mercado, inferencia de relaciones entre los datos, etc..) en general podríamos definir la minería de datos como el proceso de descubrir, extraer y almacenar información (no trivial) relevante de amplios conjuntos de datos a través de técnicas que permiten lidiar con la alta dimensionalidad de los mismos [Ángeles et al., 1998]. Sus objetivos son diversos como lo son sus áreas de aplicabilidad, estando ya inserto en muchos de los casos de forma transparente en todo tipo de software que contemple tareas como el descubrimiento de patrones de comportamiento, encontrar interrelaciones sistemáticas entre variables, construir modelos predictivos, y la extracción de información no evidente a partir de un *dataset* de datos recogidos *en bruto*.

En el proceso de KDD, además se considera la preparación de los datos, la selección y limpieza de estos, la incorporación de conocimiento previo, y la propia interpretación de los resultados provenientes de la Minería de Datos [Riquelme et al., 2006].

## 2. DIFUSIÓN Y CAMPOS DE APLICACIÓN DE LAS TÉCNICAS DE MINERÍA DE DATOS EN CIENCIA E INGENIERÍA

Se pueden citar ejemplos de aplicación exitosa de la MD en prácticamente la totalidad de las áreas de la ingeniería o la investigación científica: Física, Ciencias de la Salud, Geología, Ciencias ambientales, Meteorología, Diseño de Materiales, Sociología, Astrofísica, Optimización de diseños, por citar algunos). Cada uno de estos campos genera ingentes cantidades de datos que, en muchos casos, tardan años en ser procesados. Como ejemplo de actualidad consideremos la misión europea GAIA [1] que fue lanzada con éxito el 19 de diciembre de 2013 por la Agencia Espacial Europea desde el Puerto espacial de Kourou en la Guayana Francesa. La sonda GAIA ha supuesto un nuevo e importante desafío para probar nuestra capacidad tecnológica para tratar y procesar grandes cantidades de datos. Durante los 5 años de su misión principal, la sonda GAIA permanecerá en una órbita de Lissajous (alrededor del punto L2) del sistema Tierra-Sol, desde donde confeccionará un catálogo de más de un billón [2] de estrellas hasta magnitud 20, realizando más de 70 medidas de cada estrella, lo que permitirá obtener medidas fotométricas (que servirán para estimar muchas de sus propiedades físicas, como su luminosidad, el tipo de estrella, su fase dentro de su secuencia, su temperatura superficial y la composición de su atmosfera) determinar su posición, distancia y movimiento (por ejemplo su velocidad angular) con una increíble precisión de hasta 20 microsegundos de arco. Durante este tiempo GAIA estará disponible para enviar datos unas ocho horas todos los días que serán recibidos por la red espacio profundo a una tasa de unos 5 Mbit/s. En total GAIA enviará unos 60 TB de datos que una vez descomprimidos supondrán un total de 200 TB (terabytes) de datos en bruto.

Otro ejemplo destacable lo encontramos en el proyecto ENCODE (acrónimo del inglés ENcyclopedia Of Dna Elements) un proyecto internacional para el análisis exhaustivo del genoma humano. Desde 2003, el Proyecto ENCODE intenta dilucidar los entresijos del ADN y crear un catálogo con todos los elementos funcionales que contiene el genoma, incluyendo las partes oscuras que no codifican genes, sino que alteran el comportamiento funcional del ADN. A día de hoy, ENCODE ha recolectado tantos elementos que, si se imprimiesen sobre un mural, este mediría hasta 16 metros de alto y 30 kilómetros de largo, y que, en términos de capacidad, suman cerca de 20 TB (terabytes) de información en bruto que están disponibles públicamente en internet. Para ejemplificar la importancia del proyecto ENCODE baste decir que los datos aportados ya por ENCODE son suficientes para hacer pensar que los genes son más complejos de lo que se pensaba hasta ahora: en vez de la visión tradicional, según la cual un gen da lugar a uno o varios transcritos alternativos que codifican una proteína en sus varias isoformas, parece claro, a la luz de los datos, que una región genómica puede codificar distintos productos proteicos y además dar lugar a otros transcritos (no necesariamente codificantes de proteínas) en ambas cadenas. Todo esto ha llevado a replantear el concepto de gen, que en la era post-ENCODE se definiría como "la unión de las secuencias genómicas que codifican un conjunto coherente de productos funcionales, que son potencialmente solapantes" [Morales et al., 2013]. Es desde todo punto de vista imposible hacer aquí una enumeración completa de todas las aplicaciones que se pueden derivar del conocimiento de la secuencia genómica de organismos, pero no es difícil imaginar su impacto inmediato en campos como la salud humana (diagnóstico, tratamiento y prevención de enfermedades como el cáncer, terapia génica, farmacogenética, etc...); la mejora genética animal y vegetal; los estudios filogenéticos, de base evolutiva o poblacionales; la genética forense; la detección de especies y patógenos; la genética medioambiental, y muchas más.

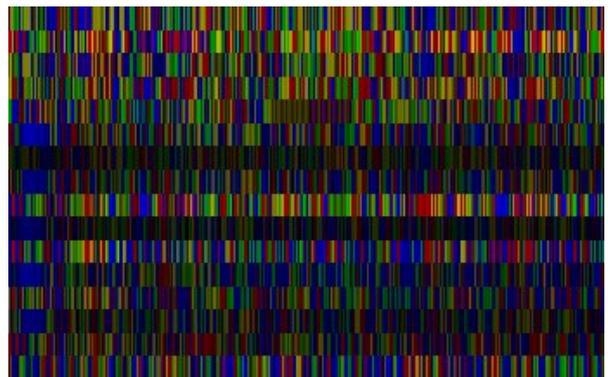

**Figura 1. Fragmento de secuenciación automática fluorescente obtenida en un secuenciador automático ABI 3130XL (Rodríguez-Tarduchy 2009).**

2.1 UN VISTAZO RÁPIDO A LAS PRINCIPALES FAMILIAS DE ALGORÍTMOS PARA MD.

Las técnicas de MD utilizan métodos para tratar la alta dimensionalidad de los datos conjuntamente a algoritmos pertenecientes al ámbito de la inteligencia artificial, así como métodos matemáticos y estadísticos que juntos permiten poder

---

(1) Se puede consultar el dosier de prensa del lanzamiento de la misión Gaia en el website de la Agencia Espacial Europea en url http://www.esa.int/esl/ESA_in_your_country/Spain/Dosier_de_Prensa_-_Mision_Gaia_-

(2) el término original en inglés (americano) es "billion", este fue traducido al español "mil millones", debido a que en español (según la Real Academia Española de la Lengua), "1 billón" equivale a "un millón de millones".

MAS NOTAS AL FINAL DEL DOCUMENTO.

realizar búsquedas de patrones, secuencias o comportamientos sistemáticos que pongan de manifiesto interrelaciones entre los datos o que sirvan para predecir comportamientos futuros [Caridad, 2001]. Estas técnicas son muy variadas, pues no todas son aplicables en cualquier conjunto de datos ni a todo procedimiento de extracción de información. En general, cualquiera que sea el problema a resolver, podemos decir que no existe una única técnica para solucionarlo y posiblemente el abanico de técnicas que comprende el campo de la MD nos permita hacer visibles diferentes realidades de nuestro conjunto de datos.

Según Aluja [Aluja, 2001], entre las técnicas más utilizadas en las labores de MD podemos encontrar:

Análisis Factoriales Descriptivos: abordan el problema de cómo analizar eficazmente la estructura de las interrelaciones (correlaciones) entre un gran número de variables con la definición de dimensiones subyacentes comunes, conocidas como factores. Utiliza las técnicas de análisis de componentes principales, análisis de correspondencias, análisis factorial. Permite analizar la estructura de los datos y proporcionan herramientas de visualización, las cuales permiten observar de mejor manera realidades multivariantes complejas y, por ende, manifestar las regularidades estadísticas.

Análisis de la Cesta de la Compra (conocidas también como *Market Basket Analysis*): usualmente se usan reglas de asociación y de secuenciación donde se analizan los datos para descubrir reglas que identifiquen patrones o comportamientos utilizando algoritmos computacionalmente intensivos. Permite detectar qué productos se adquieren conjuntamente, permite incorporar variables técnicas que ayudan en la interpretación, como el día de la semana, localización, forma de pago. También puede aplicarse en contextos diferentes del de las grandes superficies, en particular el e-comercio, e incorporar el factor temporal.

Técnicas de Agrupamiento (llamadas comúnmente de *Clustering*): son técnicas que parten de una medida de proximidad entre individuos y a partir de ahí, buscan los grupos de individuos más parecidos entre sí, según una serie de variables mesuradas. Algunos ejemplos de este grupo pueden ser el algoritmo de las k-medias o el algoritmo de clasificación jerárquica. Agrupan individuos o variables en clases que muestran un comportamiento homogéneo y, por lo tanto, permiten descubrir patrones de comportamiento.

Series Temporales: corresponde a un conjunto de observaciones realizadas respecto a un variable en momentos equidistantes de tiempo. A partir estas series que recogen el comportamiento histórico, permite modelar las componentes básicas de la serie, tendencia, ciclo y estacionalidad y así poder hacer predicciones para el futuro.

Redes Bayesianas: modelan un fenómeno mediante un conjunto de variables y las relaciones de dependencia entre ellas. Dado este modelo, se puede hacer inferencia bayesiana; es decir, estimar la probabilidad posterior de las variables no conocidas, en base a las variables conocidas; así, consiste en representar todos los posibles sucesos en que estamos interesados mediante un grafo de probabilidades condicionales de transición entre sucesos. Puede codificarse a partir del conocimiento de un experto o puede ser inferido a partir de los datos. Permite establecer relaciones causales y efectuar predicciones.

Modelos Lineales Generalizados: son modelos que permiten tratar diferentes tipos de variables de respuesta. Al mismo tiempo, los modelos estadísticos se enriquecen cada vez más y se hacen más flexibles y adaptativos, permitiendo abordar problemas cada vez más complejos.

Previsión Local: la idea de base es que individuos parecidos tendrán comportamientos similares respecto de una cierta variable de respuesta. La técnica consiste en situar los individuos en un espacio euclídeo y hacer predicciones de su comportamiento a partir del comportamiento observado en sus vecinos.

Redes Neuronales: inspiradas en el funcionamiento de la neurona biológica y en lo que conocemos del funcionamiento de la sinapsis humana, son generalizaciones de modelos estadísticos clásicos. Las RRNN son un paradigma de aprendizaje y procesamiento automático. Trata de un sistema de interconexión de neuronas que colaboran entre sí para producir un estímulo de salida. Su novedad radica en el aprendizaje secuencial, el hecho de utilizar transformaciones de las variables originales para la predicción y la no linealidad del modelo. Permite aprender en contextos difíciles, sin precisar la formulación de un modelo concreto. Su principal inconveniente es que para el usuario son una caja negra.

Árboles de Decisión: son representaciones gráficas y analíticas de datos ya establecidos mediante una base de datos. Permiten obtener de forma visual las reglas de decisión bajo las cuales operan las variables y parámetros, a partir de datos históricos almacenados. Su principal ventaja es la facilidad de interpretación. Ayuda en la toma de decisiones, desde un punto de vista probabilístico, con el fin de obtener la opción que mejor convenga.

Algoritmos Genéticos: Los algoritmos evolutivos son estrategias de optimización y búsqueda de soluciones que toman como inspiración la evolución en distintos sistemas biológicos. La idea fundamental de estos algoritmos es mantener un conjunto de individuos que representan una posible solución del problema. Estos individuos interactúan, tanto a nivel de individuo como a nivel de población y compiten, siguiendo el principio de selección natural por el cual sólo los mejor adaptados sobreviven al paso del tiempo. Esto redunda en una evolución hacia soluciones cada vez más aptas. Los algoritmos evolutivos son una familia de métodos de optimización, y como tal, tratan de hallar una tupla de valores $(x_i..., x_n)$ tales que se minimice una determinada función $F(x_i, ..., x_n)$. En un algoritmo evolutivo, tras parametrizar el problema en una serie de variables, $(x_i..., x_n)$ se codifican en una población de cromosomas. Sobre esta población se aplican uno o varios operadores genéticos y se fuerza una presión selectiva (los operadores utilizados se aplicarán sobre estos cromosomas, o sobre poblaciones de ellos). Esta forma de funcionamiento les confiere su característica más destacable: un algoritmo evolutivo puede ser implementado con un escaso conocimiento del espacio de soluciones, o a lo sumo, con un conocimiento básico de éste, siempre que tengamos un modelo capaz de predecir el comportamiento que sería esperado de una propuesta a solución, pues el conocimiento es inferido conforme avanza la exploración. Esto los hace algoritmos robustos, por ser útil para cualquier problema de optimización, pero a la vez débiles, pues no están especializados en ningún problema concreto siendo los operadores genéticos empleados los que en gran parte confieren la especificabilidad al método empleado [García-Gutiérrez & Díaz,2014].

A pesar de no ser incluida explícitamente por Aluja [Aluja, 2001], la mayoría de la literatura, hoy en día, contiene la técnica de regresión dentro de las clasificaciones.

Técnica de Regresión: es una técnica de tipo predictiva. Numerosos problemas pueden resolverse usando regresión

---

(1) Se puede consultar el dosier de prensa del lanzamiento de la misión Gaia en el website de la Agencia Espacial Europea en url http://www.esa.int/esl/ESA_in_your_country/Spain/Dosier_de_Prensa_-_Mision_Gaia_-

(2) el término original en inglés (americano) es "billion", este fue traducido al español "mil millones", debido a que en español (según la Real Academia Española de la Lengua), "1 billón" equivale a "un millón de millones".

MAS NOTAS AL FINAL DEL DOCUMENTO.

lineal. La regresión lineal es la más utilizada para formar relaciones entre datos. Es una técnica rápida y eficaz pero insuficiente en espacios multidimensionales donde puedan relacionarse más de dos variables. Sin embargo, en la naturaleza, la mayoría de los problemas presentan relaciones no lineales, es decir los datos no muestran una dependencia lineal, por lo que es necesario usar técnicas de regresión no lineal para obtener resultados ajustados a la realidad de los comportamientos. También en ocasiones pueden aplicarse transformaciones a las variables para que un problema no lineal pueda convertirse a uno lineal. Los modelos lineales generalizados representan el fundamento teórico en que la regresión lineal puede aplicarse para modelar las categorías de las variables dependientes.

Por otra parte, Caridad [Caridad, 2001] además incluye las siguientes técnicas:

Lógica Difusa: permiten manejar datos en los cuales existe una transición suave entre categorías distintas, por lo que algunos datos pueden tener propiedades de clases diferentes, estando parcialmente en más de un grupo con un grado específico de pertenencia [Rojas et al., 2008]. Esta técnica es empleada por numerosos modelos y productos industriales para representar situaciones en las que los problemas de clasificación están afectados de incertidumbre.

Según [Weiss & Indurkhya, 1998], los algoritmos de minería de datos se clasifican en dos grandes categorías: supervisados o predictivos, y no supervisados o de descubrimiento del conocimiento. También es usual el uso de sistemas expertos que permiten validar los resultados teóricos con aquellos obtenidos de forma empírica. Tradicionalmente, los sistemas expertos han sido usados para resolver procesos complejos, en los cuales hay muchos factores involucrados, y por tanto se necesita tener en cuenta una amplia base de datos históricos, sobre los que se intuye alguna regla que permita la toma de decisiones rápida. Por lo que combinar las técnicas de minería de datos con sistemas expertos se ha convertido en una línea de investigación de gran auge hoy en día.

Otra de las nuevas vías de investigación es el *fuzzy mining*, esto es, la utilización de las técnicas de minería de datos con objetos simbólicos, que representen más fidedignamente la incertidumbre que se tiene de los objetos que se estudian.

Los algoritmos supervisados o predictivos pronostican el valor de un atributo de algún conjunto de datos, conocidos ya otros atributos. A partir de datos cuyo atributo se conoce se puede inducir una relación entre dicho atributo y otra serie de ellos (atributos). Esas relaciones sirven para realizar la predicción en datos donde el atributo es desconocido. Este proceso se conoce como aprendizaje supervisado y se desarrolla en dos fases: una primera fase de entrenamiento, donde se construye un modelo usando un subconjunto de datos con atributo conocido, y otra fase de prueba, la cual consiste en evaluar el modelo sobre el resto de los datos [Moreno et al., 2001]. Por ejemplo, entre los algoritmos supervisados o predictivos podemos nombrar las técnicas de: árboles de decisión, inducción neuronal, regresión y series temporales.

Existen otras ocasiones donde una aplicación no tiene el potencial necesario o no aporta suficiente información a priori, para dar una solución predictiva. En ese caso hay que recurrir a los métodos no supervisados o de descubrimiento del conocimiento que revelan patrones y tendencias en los datos actuales, sin utilizar datos históricos. Entre estos están: la detección de desviaciones, segmentación, agrupamiento (clustering), reglas de asociación y patrones secuenciales [Moreno et al., 2001].

2.2 LA MINERIA DE DATOS EN CIENCIA E INGENIERÍA.

Debido al gran y rápido crecimiento de las redes de interconexión de sistemas digitales y al incremento del ancho de banda que ha permitido enviar gran cantidad de datos entre distintos sistemas digitales, la Minería de Datos se encuentra ya presente en muchas áreas de la vida cotidiana y en el segmento industrial y tecnológico, donde constituye ya un caso de éxito. Prácticamente cualquier dispositivo que genere un flujo de información proveniente de sensores de cualquier naturaleza es una fuente potencial de datos, los cuales quedan almacenados digitalmente. Básicamente, si abordamos un problema sobre un dataset sobre el que existen datos históricos disponibles desde una o varias fuentes estamos ante un problema susceptible de ser tratado mediante técnicas de minería de datos, y la importancia de esta en su resolución será directamente proporcional a la complejidad del conjunto de datos analizado.

Sea cual sea el sector al que nos refiramos con total seguridad encontraremos ejemplos del uso de una o varias técnicas de minería de datos bien sea de forma implícita o explícita, consideramos que algunos de los más importantes son: el análisis y fragmentación del mercado (marketing), la predicción bursátil y gestión de finanzas, el control de fraudes u operaciones irregulares, la seguridad y auditoría informática, la optimización de procesos industriales, las diferentes ramas de la ingeniería, el análisis de tráfico y saturación de redes, las telecomunicaciones, las ciencias de la salud, provenientes de las diferentes ramas de la Bioinformática (secuenciación, plegado de proteínas, diseño de fármacos, búsqueda de marcadores genéticos, etc..) y la biotecnología, las distintas aplicaciones en ciencias físicas e astrofísica (procesamiento de señales, detección de patrones, navegación autónoma, calibración de instrumentos, validación o parametrización de modelos numéricos, análisis de datos heterogéneos o de alta dimensionalidad, sistemas de control y respuesta adaptativa, visión artificial y detección de formas y obstáculos, optimización diseños o de materiales, corrección automática de ruido, etc..), sistemas de aviso preventivo de fraudes, clasificación de riesgos en seguros, pólizas o de retorno de inversión, modelización de procesos, educación adaptativa, y podríamos seguir citando muchas más.

En el campo de la ingeniería en general se colecta y almacena gran cantidad de datos a través de diferentes sensores (mediciones de diferentes variables), donde la MD juega un rol importante para la creación de modelos y patrones. En computación, las técnicas de Minería de Datos se utilizan para la obtención de modelos del usuario e interfaces adaptativas, para hallar patrones en diferentes formatos de los datos (por ejemplo, en los análisis heurísticos para la protección antivirus), para el análisis de saturación en redes de computadores, para la indexación de contenido y en general, para la extracción de información valiosa, por ejemplo en la construcción y validación de modelos de ingeniería del software [Moreno et al., 2001]. En Ingeniería Civil se miden y almacenan datos provenientes de sensores para monitorizar y analizar las condiciones de las estructuras o para predecir las condiciones de ruptura o agrietamiento en estructuras de

---

(1) Se puede consultar el dosier de prensa del lanzamiento de la misión Gaia en el website de la Agencia Espacial Europea en url http://www.esa.int/esl/ESA_in_your_country/Spain/Dosier_de_Prensa_-_Mision_Gaia_-

(2) el término original en inglés (americano) es "billion", este fue traducido al español "mil millones", debido a que en español (según la Real Academia Española de la Lengua), "1 billón" equivale a "un millón de millones".

MAS NOTAS AL FINAL DEL DOCUMENTO.

hormigón. En ingeniería eléctrica, las técnicas de MD son utilizadas para monitorizar las condiciones de las instalaciones de alta tensión. En el área de ingeniería de transporte la MD es muy útil para el análisis y toma de decisiones para generar vías óptimas o inteligentes de transporte.

También el diagnóstico de fallos en cualquier proceso es un área de investigación de gran auge hoy en día, por lo que en [Ríos-Bolívar et al., 2014] presentan aplicaciones de diagnóstico de fallas a partir de Minería de Datos a través de índices estadísticos y coeficientes polinomiales como una nueva técnica para abordar estos problemas. En el campo de la multimedia [Wlodarczak et al., 2015] para la búsqueda e identificación de imágenes, video, reconocimiento de voz, búsqueda texto en grandes cantidades de repositorios de datos. Las innovaciones tecnológicas en el mundo audiovisual producen el almacenamiento de una cantidad enorme de datos. El análisis de estos datos permite una mejora en la toma de decisiones por parte de las organizaciones implicadas [Aluja, 2001].

En telecomunicaciones, el rápido desarrollo de la tecnología y la mercadotecnia ha producido un crecimiento sorprendente en la demanda por el uso de MD. Por ejemplo, las técnicas de MD son demandadas y ampliamente utilizadas hoy día en el análisis multidimensional de datos de conmutación en telecomunicaciones (tratamiento de datos como tiempo de llamada, tipo de llamada, ubicación del usuario, duración de la llamada, etc.). Estos datos almacenados pueden ser usados a través de MD para identificar y comparar el comportamiento de los usuarios, su perfil, uso de recursos, etc... [Han & Kamber, 2006]. También en este campo es usada la MD para la identificación de fraudes, actuando tanto de manera preventiva (identificar potenciales usuarios que puedan cometer algún fraude) como identificar fraudes cometidos y que se hayan pasado por alto.

En el área de las ciencias de la salud, se utilizan para la detección precoz y prevención de enfermedades, para el análisis de marcadores genéticos, para prever la probabilidad de una respuesta satisfactoria a un tratamiento médico, como por ejemplo las reglas de asociación utilizadas en [Marchán et al., 2011] para determinar factores de riesgo epidemiológico de transmisión de enfermedades, para asistir al médico en el diagnóstico, por ejemplo detectar patrones anormales en los análisis bioquímicos o en las pruebas de imagen y diagnóstico digital. En el área farmacéutica para identificar nuevos fármacos, moléculas y tratamientos (diseño de moléculas y sustancias con acción farmacológica), e incluso en la investigación forense para la identificación de restos humanos [Medina et al., 2012].

Este campo de aplicación es de hecho tan grande y se hace tan necesaria la minería de datos (los problemas en investigación biomédica requieren a menudo de análisis multidimensional y computación de alto rendimiento) que ha dado lugar a una nueva rama ciencia, la bioinformática. Por ejemplo, para la identificación de secuencias de ADN, se requiere de un alto nivel computacional, estadístico, de programación matemática, y minería de datos para desarrollar una estrategia de búsqueda e identificación efectiva. Entre otras aplicaciones en esta área podemos nombrar [Khalid, 2010] la predicción de la estructura de proteínas, clasificación de genes, asistencia a la detección de diferentes tipos cáncer, modelos estadísticos para iteraciones entre proteínas, diagnóstico precoz de enfermedades, asistencia al tratamiento y seguimiento clínico, modelización de procesos bioquímicos, entre otros [Medina et al., 2012].

En meteorología y ciencias ambientales se utilizan las técnicas de MD en ámbitos tan distintos como: la predicción de tormentas, modelización y predicción de terremotos, reconstrucciones geológicas, detección de incendios, seguimiento de glaciares, control de la contaminación humana o la actividad industrial, distribución y mejora de cultivos o para el tratamiento de plagas y enfermedades [Medina et al., 2012]. o para identificar si el epicentro de un terremoto puede ser agrupado dentro de una falla continental especifica. También es usada para el análisis y predicción de desastres naturales como lo expone [Refonaa et al., 2015].

En el campo de la seguridad encontramos aplicaciones para reconocimiento facial o dactilar, identificación de personas, o biométrica, seguimiento de operaciones bancarias, investigación terrorista, rastreo de personas, seguimiento de operaciones bancarias, etc... [Deepa et al., 2013]. Inclusive en el área judicial, donde se almacena gran cantidad de datos sobre los juicios, la MD de datos tiene un rol en la búsqueda de patrones para identificar casos similares.

En el campo de la educación recientemente ha cobrado importancia la llamada Minería de Datos Educacional (del inglés *EDM por Educational Data Mining*), el cual trata del desarrollo de métodos para descubrir conocimiento (donde se apoya en la rama de computación semántica) de datos almacenados en ambientes educacionales. Sus objetivos son identificar y predecir el comportamiento futuro de los estudiantes, estudiar los efectos del soporte pedagógico, predecir los resultados de los estudiantes, etc... [Rajkumar, 2014]. En base a estos estudios y resultados, las instituciones educativas pueden enfocarse en qué enseñar y cómo hacerlo [Lakshmi & Shanavas 2014, Peña-Ayala 2014] realizaron un trabajo de recopilación y análisis sobre los trabajos y aplicaciones más recientes para la fecha en minería de datos educacional, seleccionando 240 artículos relacionados con este nuevo campo, descubriendo la existencia de 222 diferentes métodos en minería de datos educacional y 18 tipos de técnicas aplicadas.

Otro campo que también está tomando mucho interés en años recientes es denominado (PPDM, Privacy Preserving Data Mining). La idea de PPDM es modificar los datos de tal manera que se pueda aplicar algoritmos de MD eficientemente sin comprometer la información sensible que se encuentre en los datos almacenados [Xu et al., 2014].

Tal y como lo predijo Aluja (Aluja, 2001), el campo de actuación de la Minería de Datos no puede sino crecer. En particular debemos mencionar en estos momentos el análisis de datos en tiempo real mediante las diferentes redes interconectadas obtenidos en línea, dando lugar al *Web Data Mining*, donde las técnicas de MD se utilizan para analizar flujos de datos conforme estos se producen. Recientemente [Aggarwal, 2015] expresa que el número de documentos indexados en la web alcanza el orden de los mil millones, siendo el número de documentos no visibles directamente aún mucho mayor. Sin duda la aparición y popularización de las aplicaciones de Web Semántica (contenido estructurado y contextualizado) supondrá una revolución en este ámbito en los próximos años.

---

(1) Se puede consultar el dosier de prensa del lanzamiento de la misión Gaia en el website de la Agencia Espacial Europea en url http://www.esa.int/esl/ESA_in_your_country/Spain/Dosier_de_Prensa_-_Mision_Gaia_-

(2) el término original en inglés (americano) es "billion", este fue traducido al español "mil millones", debido a que en español (según la Real Academia Española de la Lengua), "1 billón" equivale a "un millón de millones".

## 3. ABORDAJE DE LA ESCALABILIDAD EN ALGORITMOS DE MINERÍA DE DATOS

El término escalabilidad cuando se refiere a algoritmos de Minería de Datos (MD) nos indica la capacidad intrínseca del método para manejar el crecimiento en el tamaño de la entrada sin que con ello se degrade la calidad de la solución encontrada ni se produzca una degradación en el rendimiento del algoritmo o esta se produzca, pero no hagan al método computacionalmente inmanejable. Dado el volumen de datos a tratar, el coste de los algoritmos ha de ser todo lo lineal que sea posible respecto de los parámetros que definen el coste, en particular respecto del número de atributos [Aluja, 2001]. Su importancia surge debido a que el espacio de búsqueda es usualmente de orden exponencial en cuanto al número de atributos se refiere. La mayoría de los algoritmos utilizados en el proceso de MD son de naturaleza iterativa, requiriendo que las bases de datos sean revisadas (escaneadas) varias veces, lo que lógicamente hace el método bastante pesado y costoso computacionalmente hablando. El uso de muestras puede muchas veces ser sensible a la tendencia de los datos, y esto puede afectar el desempeño del algoritmo utilizado [Mitra & Acharya, 2003]. Debido a que los problemas que se tratan involucran un gran número de variables y de datos, la eficiencia computacional y la escalabilidad son temas de gran importancia para la aplicación de estos métodos. Por ejemplo, técnicas desarrolladas dentro del enfoque estadístico, normalmente involucran operaciones computacionalmente costosas (p.e. inversión de matrices). En consecuencia, la aplicación directa de estos métodos sobre grandes conjuntos de datos, en muchos casos resulta inoperante y requerirán de una fase inicial de filtrado y selección de atributos que maximizar la representabilidad del subconjunto de entrada con la menor afección en el rendimiento.

Para que un algoritmo sea escalable, su tiempo de ejecución debe crecer aproximadamente de forma lineal en proporción al tamaño de los datos, dados los requerimientos y recursos del sistema como lo son: suficiente espacio en la memoria principal y en el disco duro.

Hablar de escalabilidad en algoritmos de MD está muy relacionado con hablar sobre la eficiencia de los mismos para el procesamiento de los grandes almacenamientos de datos. Desde el punto de vista del descubrimiento del conocimiento, eficiencia y escalabilidad son términos claves en la implementación de sistemas de MD.

En Minería de Datos existen dos tipos de problemas comunes para la escalabilidad, está la escalabilidad de filas (se refiere al número de instancias existentes o de forma más general al tamaño del almacén de datos) y la de columnas (se refiere al número de atributos o características estudiadas, y por tanto influye directamente en el tamaño y dimensiones del espacio de búsqueda). Se dice que un sistema de Minería de Datos se considera escalable en filas sí, cuando el número de filas es agrandado, por ejemplo, diez veces, no toma más de diez veces realizar su ejecución. Por otra parte, un sistema de MD es considerado escalable en columnas sí, el tiempo de ejecución se incrementa linealmente respecto al número de atributos (columnas o dimensiones), siendo un mayor reto el escalamiento de columnas que el de filas [Han & Kamber, 2006].

Según Riquelme [Riquelme et al., 2006], las principales metodologías para la escalabilidad en algoritmos de Minería de Datos incluyen diseñar algoritmos rápidos y segmentar los datos. La primera se refiere a la mejora (reducción) de la complejidad computacional, tratando de optimizar la búsqueda o encontrar soluciones próximas a problemas computacionales complejos. Segmentar los datos consiste en dividir o agrupar los datos en subconjuntos (según sus instancias o características), aprender de uno o más de los subconjuntos seleccionados, y posiblemente combinar los resultados.

La propia naturaleza del método influye de forma extraordinaria en la capacidad del mismo para comportarse bien ante cambios bruscos en el tamaño de la entrada y conservar buenos rendimientos aunque la complejidad del problema aumente considerablemente, por ejemplo, los algoritmos genéticos son intrínsecamente paralelos por su forma de operar basada en población, de la misma forma, un árbol de decisión maneja de forma extraordinaria el cambio en el número de atributos de entrada puesto que parte del trabajo realizado por el algoritmo en la ejecución previa pueden ser reutilizados. De forma más general, para las diferentes familias de algoritmos utilizados con más frecuencia en tareas de minería de datos podemos decir con respecto a su escalabilidad que:

Redes Neuronales: son algoritmos desarrollados en el ámbito de la inteligencia artificial. Su algoritmo se basa en reproducir el comportamiento de nuestro cerebro, en cuanto a aprender del pasado y aplicar el conocimiento aprendido a la resolución de nuevos problemas mediante la formación conceptual. Las redes neuronales pueden servir para predicción, clasificación, segmentación e incluso para aprendizaje no supervisado. Estas permiten determinar reglas de clasificación de tipo no paramétrica, que proporcionan mejores resultados cuando la separación entre clases es no lineal. Las redes neuronales difieren fundamentalmente respecto a las técnicas tradicionales, en que estas son capaces de detectar y aprender complejos patrones y características dentro del conjunto de datos [García & Molina., 2012]. La principal ventaja se observa en que una red neuronal aprende, como resultado del entrenamiento, lo cual le permite adaptarse a nuevos cambios o dinámicas que se presenten. En casos de muy alta complejidad, la mayoría de los algoritmos escalables descritos anteriormente resultan fallidos, siendo las redes neuronales una alternativa bastante satisfactoria. Una característica innovadora sobre los algoritmos tradicionales es que son capaces de trabajar con datos incompletos.

Lógica Difusa: existen casos en los que los datos no pueden tratarse de manera determinista, es decir su agrupamiento es difuso debido a que pueden pertenecer, con cierto grado, a diferentes grupos. Los algoritmos basados en lógica difusa surgen de esta necesidad, permitiendo tratar datos en los cuales existe una transición suave entre categorías (o clases) distintas, por lo que algunos datos pueden tener propiedades de clases diferentes, estando parcialmente en más de un grupo

---

(1) Se puede consultar el dosier de prensa del lanzamiento de la misión Gaia en el website de la Agencia Espacial Europea en url http://www.esa.int/esl/ESA_in_your_country/Spain/Dosier_de_Prensa_-_Mision_Gaia_-

(2) el término original en inglés (americano) es "billion", este fue traducido al español "mil millones", debido a que en español (según la Real Academia Española de la Lengua), "1 billón" equivale a "un millón de millones".

MAS NOTAS AL FINAL DEL DOCUMENTO.

con un grado específico de pertenencia [Ghosh et al., 2014] .En general, las reglas difusas combinan uno o más conjuntos difusos de entrada, llamados antecedentes o premisas, y les asocian un conjunto difuso de salida, llamado consecuente o consecuencia. La complejidad de estos algoritmos es mayor comparada a los algoritmos tradicionales, debido a que pasamos de valores deterministas a funciones de posibilidad para definir los grados de pertenencia de los datos. El concepto de lógica difusa es combinado comúnmente con algoritmos de *clustering* para mejorar la característica de escalabilidad de estos últimos. En principio, el modificar los algoritmos de *clustering* con técnicas de lógica difusa arrojaba buenos resultados en cuanto a eficiencia y escalabilidad se trata para bases de datos pequeñas o hasta medianas. Sin embargo, para grandes cantidades de datos otras técnicas (como las basadas en redes neuronales) arrojaban mejores resultados, debido principalmente a que los tiempos computacionales al usar lógica difusa no eran prácticos. Es entonces cuando se comienza a trabajar con particiones de datos que reflejen en su mayoría a todo el gran conjunto de datos, reduciendo así los tiempos computacionales. Sin embargo, el usar lógica difusa en grades bases de datos aun es un reto, requiriendo de estrategias, estructuras y algoritmos más sofisticados, como por ejemplo el uso de computación paralela. La selección de los algoritmos basados en lógica difusa requiere la consideración de un balance entre la calidad del algoritmo y la velocidad de ejecución del mismo [Mathew & Vijayakumar, 2014]

Los Algoritmos Genéticos (AG): al igual que las redes neuronales, tienen sus fundamentos conceptuales en la biología, en este caso tratando de emular los mecanismos genéticos de los organismos biológicos; es decir, en la idea de que los individuos en una población compiten con cada uno de los otros por los recursos, la supervivencia de los mejor adaptados. Son métodos adaptativos que pueden usarse para resolver problemas de búsqueda y optimización, por medio de un proceso evolutivo artificial. Se caracteriza por su definición de poblaciones de individuos, asignándole un valor a cada individuo, y cada individuo representa una solución factible a un problema dado. Su principal campo de aplicación abarca mayormente problemas que no encuentren soluciones aceptables u óptimas (dentro de cierto criterio) usando algunas técnicas especializadas para dicho problema, en muchos otros casos, los algoritmos especializados pueden combinarse con algún algoritmo genético y ampliar sus potencialidades. Una de sus principales ventajas computacionales es el permitir que sus operaciones se puedan ejecutar en paralelo, pudiendo distribuir las tareas en diferentes procesadores. El desempeño de un AG depende mucho del método que se seleccione para codificar las soluciones candidatas, de los operadores y de la fijación de los parámetros. Se ha establecido que la eficiencia de los AG para encontrar una solución óptima está determinada por el tamaño de la población. Así, una población grande requiere de más memoria para ser almacenada. [García & Molina., 2012].

Los árboles de decisión: son una técnica de clasificación basada en reglas *If-then-else* (sí-entonces-sino). Los árboles de decisión dividen un espacio de decisión en regiones constantes a trozos, siendo los algoritmos más usados como métodos de aprendizaje para la exploración de datos. Son fáciles de interpretar y no requieren información *a-priori* (previa) sobre la distribución de los datos [Mitra & Acharya, 2003]. El concepto fue popularizado por [Quinlan y col., 1986] con la propuesta del algoritmo ID3 (*Interactive DiChaudomizer*3), dando a conocer que una de las principales ventajas de los árboles de decisión es su capacidad de dividir una decisión (o problema) compleja en una colección de simples decisiones (o sub-problemas) (Mitra y col., 2003). Basados en el algoritmo ID3, otros algoritmos como por ejemplo C4.5 [Quinlan & Ross, 1993], C5.0 y los denominados CART (*Classification and Regression Trees*) fueron propuestos, incrementando las ventajas de ID3. La eficiencia de estos algoritmos (ID3, C4.5, C5.0, y CART) fue considerada bastante buena para pequeños grupos de datos [Hssina et al., 2014], sin embargo, los resultados no fueron muy satisfactorios para aplicaciones en el proceso de Minería de Datos de grandes cantidades de datos. Surge así la necesidad, dentro de los algoritmos basados en arboles de decisión, de desarrollar algoritmos más avanzados que sean escalables a grandes conjuntos de datos, entre ellos están: SLIQ (*Supervised Learning In Quest*), SPRIN (*Scalable PaRallelizable INndution of decision Trees*) y RainForest [Han & Kamber, 2006].

En cuanto a los algoritmos de *clustering*, uno de los primeros algoritmos de búsqueda clásico en labores de agrupamiento el algoritmo k-medias [3], ha sufrido algunas modificaciones y revisiones posteriores que le permiten ampliar su potencialidad, en cuanto a escalabilidad se refiere, reteniendo atributos esenciales en la memoria principal, mientras se simplifican otros elementos pertenecientes a sub-agrupaciones y descartando datos redundantes, por lo tanto grandes cantidades de datos son modelados usando solamente el algoritmo de agrupamiento y los atributos retenidos en la memoria principal [Ghosh et al., 2014].

En los últimos quince años son muchas las propuestas de modificaciones a los algoritmos básicos de Minería de Datos que buscan mejorar el desempeño de estos cuando se enfrentan a problemas de alta dimensionalidad. Por su difusión y uso tanto en el ámbito académico como en diferentes aplicaciones comerciales podríamos destacar los siguientes:

Árbol C4.5: Los árboles de clasificación están especializados en predecir la clase de una instancia a partir del resto de atributos. Predecir la clase de un registro dado los valores del resto de los atributos. Durante la fase de construcción del árbol, la mayoría de los algoritmos de árbol para minería de datos repetidamente seleccionan una pregunta que hacer a un nodo. Una vez que la respuesta a esa pregunta es determinada, los registros de base de datos se dividen en dos según la evaluación realizada en el punto anterior. El proceso se repite en cada nodo hijo, hasta que se cumpla un criterio de parada y el nodo se convierte en una hoja que predice la clase más frecuente en esa partición. Durante el crecimiento del árbol, C4.5 utiliza un algoritmo de ganancia de información (en weka *infogainratio*) resultando en un clasificador altamente eficiente. Durante la poda, el algoritmo C4.5 pone a prueba la hipótesis de que la sustitución de un nodo con una hoja no aumentará el error del árbol. Si la hipótesis no puede ser rechazada, la poda
se produce, y el nodo se sustituye con una hoja.

---

(1) Se puede consultar el dosier de prensa del lanzamiento de la misión Gaia en el website de la Agencia Espacial Europea en url http://www.esa.int/esl/ESA_in_your_country/Spain/Dosier_de_Prensa_-_Mision_Gaia_-

(2) el término original en inglés (americano) es "billion", este fue traducido al español "mil millones", debido a que en español (según la Real Academia Española de la Lengua), "1 billón" equivale a "un millón de millones".

MAS NOTAS AL FINAL DEL DOCUMENTO.

PAM (*Partitioning Around Medoids*): es un algoritmo de agrupamiento, consiste en asumir la existencia de N objetos, así para encontrar los k-agrupamientos (k-clusters) se debe determinar un objeto que sea representativo en cada agrupación. El objeto representativo es un objeto centralmente localizado dentro del grupo (centroide), y este se denomina [2]centroide (objeto, no restringido a un valor numérico). Se deben seleccionar los [2]centroides de k para los k-agrupamientos. Así el algoritmo intenta analizar todos los pares posibles de objetos, tales que, cada objeto no seleccionado es agrupado con el centroide más similar [Santhosh et al., 2012]. Este algoritmo presenta buena efectividad, pero no es escalable para grades volúmenes de datos.

CLARA (*Clustering LARge Applications*): es un algoritmo de agrupamiento que está basado en el muestreo. Este algoritmo elige al azar una pequeña porción (muestra) del total de datos. Luego, usando el algoritmo computacional PAM[1], se seleccionan (de esta muestra): un representante de los datos y los centroides (k). La muestra seleccionada al azar es tomada como una representación correcta del conjunto total de datos, por lo tanto, los objetos representativos (centroides) elegidos serán similares como si se eligieran del conjunto total de datos. Sin embargo, en caso de que la muestra del centroide no se encuentre dentro de los mejores k-centroides, el algoritmo no puede encontrar el mejor agrupamiento (*clustering*), por lo que pierde eficiencia. Además, sí S (tamaño de la muestra) no es lo suficientemente grande, la efectividad del algoritmo es baja, y por otro lado sí S es demasiado grande el rendimiento del algoritmo no arroja buenos resultados. La complejidad computacional de este algoritmo para cada iteración es $O(kS2) + k(N-k))$, donde N es el número de objetos (Hu y col., 2013).

CLARANS (*Clustering Large Applications based on RANdomized Search*): también es un algoritmo de agrupamiento. Este algoritmo combina los algoritmos PAM y CLARA. Básicamente el proceso que realiza es encontrar una muestra con una cierta aleatoriedad en cada paso de la búsqueda. El agrupamiento obtenido después de sustituir un solo centroide se denomina el vecino del agrupamiento actual. Si se encuentra a un mejor vecino, CLARANS se mueve al nodo del vecino y el proceso comienza de nuevo; si no, el agrupamiento actual produce un grado óptimo local. Si se encuentra en el grado óptimo local, CLARANS comienza con un nuevo nodo aleatoriamente seleccionado en búsqueda de un nuevo grado óptimo local [Ng & Han, 2002]. Experimentalmente CLARANS ha demostrado ser más eficiente que PAM y CLARA. La complejidad computacional es $O(N2)$; sin embargo, el rendimiento del agrupamiento depende del método usado para escoger la muestra, y para N muy grandes no se puede garantizar el rendimiento de la muestra.

DBSCAN (D*ensity-Based Spatial Clustering of Applications with Noise*): el denominado agrupamiento espacial basado en densidad de aplicaciones con ruido es un algoritmo que pertenece a la familia de algoritmos de conglomeración espacial. Aborda la integración entre la Minería de Datos espaciales y la interfaz con el sistema de bases de datos espaciales. El algoritmo comienza por un punto arbitrario que no haya sido visitado. La vecindad de este punto es visitada, y si contiene suficientes puntos, se inicia un agrupamiento sobre el mismo. De lo contrario, el punto es etiquetado como ruido [Hu & Ester, 2013]. Una de sus ventajas es que no todos los datos deben permanecer en memoria principal. En cuanto a las complejidades computacionales, si se utiliza una estructura de índice que ejecute la consulta a los vecinos es $O(Log(N))$, la complejidad temporal total es $O(N*Log(N))$; sin usar la estructura de índice, la complejidad temporal es $O(N^2)$. En ocasiones se construye una matriz de distancias para evitar recalcular las distancias, esto necesita $O(N^2)$ de memoria, mientras que con una implementación que no se base en matrices solo se necesita $O(N)$ de memoria. En 2014, el algoritmo DBSCAN fue merecedor del premio a la prueba del tiempo [SIGKDD, 2014].

BIRCH (*Balanced Iterative Reducing and Clustering Using Hierarchies*): es un algoritmo de grupo jerárquico que genera una estructura de árbol, donde cada nodo (del árbol) representa un grupo, las hojas del árbol son aquellos grupos de menor tamaño o jerarquía, y un grupo de mayor jerarquía contiene a su subgrupo de hijos. La ventaja de BIRCH consiste en que no se guarda cada objeto en cada grupo (nodo), sino que se guarda información simplificada de los objetos que lo componen, como, por ejemplo: el centroide y el radio del grupo. Requiere revisar el conjunto de datos una sola vez, por lo que la complejidad computacional es $O(N)$.

CURE (*Clustering Using REpresentatives*): es otro algoritmo jerárquico. Trata de seleccionar objetos, de un conjunto de datos donde los objetos se encuentran bastante dispersos, y condensarlos en el centroide del grupo de acuerdo a un parámetro denominado factor de condensación. Este algoritmo mide la similitud existente entre dos grupos, a través de la similitud de las parejas de objetos más cercanos pertenecientes a grupos diferentes [Aggarwal & Reddy., 2014] .Permite evitar algunos de los problemas asociados al ruido; sin embargo, por ser un algoritmo de agrupamiento individual presenta la desventaja de que se pueden tomar decisiones incorrectas debido a la inconsistencia misma de los datos. La complejidad computacional es de $O(N)$.

IREP (*Incremental Reduced Error Pruning*): intenta abordar algunos de los problemas encontrados con una reducción de error de poda (REP) en programación lógica inductiva, mejorando notablemente su eficiencia [6]. La poda es una forma estándar de tratar con el ruido en el aprendizaje de árbol de decisión. REP introduce el concepto de post-poda, donde se genera primero una descripción conceptual que explica perfectamente todas las instancias de entrenamiento. La teoría se generaliza posteriormente cortando las ramas del árbol de decisión. Este enfoque tiene varias desventajas, que son abordados por él algoritmo. IREP integra pre y post poda como una solución alternativa. La pre-poda se traduce en que durante la formación conceptual algunos patrones de entrenamiento son ignorados deliberadamente. Después de aprender una cláusula en fase de post-poda, algunos literales pueden ser eliminados de una cláusula permitiendo un cierto grado de perdida de precisión.

CLIQUE (*Clustering In QUEst*): este algoritmo identifica automáticamente subspacios en un espacio de alta dimensión, de tal manera de que se puedan obtener mejores agrupamientos que en el espacio original. Puede interpretarse

---

(1) Se puede consultar el dosier de prensa del lanzamiento de la misión Gaia en el website de la Agencia Espacial Europea en url http://www.esa.int/esl/ESA_in_your_country/Spain/Dosier_de_Prensa_-_Mision_Gaia_-

(2) el término original en inglés (americano) es "billion", este fue traducido al español "mil millones", debido a que en español (según la Real Academia Española de la Lengua), "1 billón" equivale a "un millón de millones".

MAS NOTAS AL FINAL DEL DOCUMENTO.

como un método de agrupamiento basado en celdas (grids), ya que se divide cada dimensión en el mismo número de intervalos (de la misma longitud). A su vez, también puede interpretarse como un método de agrupamiento basado en densidad puesto que una unidad se considera densa si la fracción de puntos contenida en ella excede un parámetro del modelo [Aggarwal & Reddy, 2014].

Otros algoritmos que consideran el problema de la escalabilidad son los basados en arboles de decisión, capaces de trabajar con grandes conjuntos de datos, sin tener que mantener en memoria todo el conjunto de datos necesario para generar el árbol de decisión. Entre ellos tenemos:

SLIQ (*Supervised Learning In Quest*), es un árbol de decisión diseñado para clasificar grandes cantidades de datos. Primeramente, realiza una tarea de preclasificación en la fase de crecimiento de árbol, lo cual hace que la clasificación no sea compleja para cada nodo. La idea principal del algoritmo se basa en tener una lista por cada atributo. La principal problemática que presenta es tener que almacenar en memoria todo el conjunto de entrenamiento para construir el árbol de decisión [Aggarwal & Reddy 2014].

SPRINT (*Scalable PaRallelizable INndution of decision Trees*): es básicamente una mejora del algoritmo SLIQ, continúa con la idea de tener una lista por cada atributo. Su diferencia con SLIQ radica en la forma de cómo se representan las listas para cada atributo. Al igual que SLIQ presenta problemas de memoria [Aggarwal & Reddy 2014].

Para ensanchar la característica de escalabilidad en los algoritmos basados en árboles de decisión, un nuevo algoritmo llamado *RainForest* fue propuesto.

*RainForest*: principalmente trata de reducir las estructuras de manera que puedan ser almacenadas en memoria. Se accede al conjunto hasta dos veces en cada nivel del árbol, con lo que resulta en un mayor tiempo de procesamiento. La idea es formar conjuntos de listas denominados AVC (Atributo-Valor, Clase) para describir a los objetos del conjunto de entrenamiento. Estas listas contendrán a todos los posibles valores que un atributo pueda tomar, además de su frecuencia en el conjunto de entrenamiento. De esta manera, las listas no serán de gran longitud y con ello podrán ser almacenadas en memoria [Gama et al., 2004].

En la literatura podemos encontrar una gran variedad de algoritmos escalables, basados estos en los algoritmos descritos anteriormente. En ocasiones los nuevos algoritmos propuestos presentan alguna ligera o considerable modificación, otros presentan avance utilizando los nuevos desarrollos tecnológicos al pasar de los años, y muchos otros surgen de combinaciones con nuevas técnicas computacionales.

Una de las desventajas de los algoritmos de clustering jerárquicos es que su propiedad de escalamiento es muy pobre, y en gran probabilidad arrojan resultados fallidos. Por lo que este tipo de algoritmos han sido combinados con otras técnicas para poder tratar con grandes cantidades de datos. Por un lado, BIRCH combina los conceptos de agrupamiento y estructura de árboles para simplificar representaciones de agrupamiento, dando a lugar una mayor velocidad y propiedad de escalabilidad para grandes conjuntos de datos, haciéndolo más efectivo y dinámico a la introducción de nuevos atributos. Este algoritmo escala linealmente, encontrando un buen agrupamiento con solo una revisión (escaneo) del conjunto de datos, y además puede mejorar la calidad del mismo con revisiones adicionales; sin embargo, si es muy sensible al orden del registro de los datos. Por su parte, CURE utiliza la técnica de seleccionar del agrupamiento, objetos lo bastante esparcidos para luego comprimirlos hacia el centro del grupo y con esto mejorar su escalabilidad y eficiencia.

El algoritmo CLIQUE tiene la ventaja de encontrar automáticamente los sub-espacios de máxima dimensión en los que existen agrupamientos de alta densidad, además no depende del orden de presentación de los datos ni presupone ninguna distribución de datos. Por lo que es un algoritmo que escala linealmente respecto al tamaño de la entrada de datos y presenta buena propiedad de escalamiento en cuanto a número de dimensiones (atributos) se refiere. Sin embargo, la exactitud de los resultados del agrupamiento puede verse degradada por la simplicidad del método [Han & Kamber, 2006].

Los algoritmos basados en árboles de decisión son escalables, pudiendo procesar conjuntos de datos con independencia del número de clases, atributos y registros. No obstante, algunos algoritmos continúan presentando restricciones en cuanto al manejo de la memoria se refiere, por la representación que usan para manejar el conjunto de datos, como por ejemplo los algoritmos descritos: SLIQ y SPRINT. *RainForest* es un algoritmo escalable para la inducción de un árbol de decisión, desarrolla un algoritmo de escalamiento de clasificación bayesiana que requiere una sola revisión (escaneada) de todo el conjunto de datos en la mayoría de los casos, y este además resuelve los problemas de memoria presentados por SLIQ y SPRINT.

En cuanto al uso de nuevas técnicas y algoritmos (como algoritmos basados en redes neurales, lógica difusa, algoritmos genéticos) se puede decir que, las redes neuronales son algoritmos escalables que proporcionan buenos resultados en problemas no lineales cuando se dispone de un número elevado de datos, su eficiencia no se degrada a pesar de trabajar con un gran número de atributos de entrada. Los investigadores en algoritmos genéticos han experimentado con varios métodos de escalamiento. Para que los AG resulten menos susceptibles a una convergencia prematura, los métodos de Escalada Sigma [4] y Escalada por la Máxima Pendiente [5] [Vélez & Silva, 2000] fueron propuestos.

Por lo tanto, el éxito de los algoritmos genéticos depende de un equilibrio entre la selección de los operadores, la definición de la función de aptitud y de la codificación.

El principio de lógica difusa, debido a su complejidad y requerimientos de memoria (en ocasiones esta crece exponencialmente ante entradas lineales), ha sido combinado con diferentes algoritmos de otra naturaleza para mejorar su escalabilidad y eficiencia. Otro de los problemas que presenta el diseñar algoritmos usando lógica difusa, es el tener que construir funciones de membresía a cada entrada. Algoritmos tradicionales como CLARANS han sido combinados con

---

(1) Se puede consultar el dosier de prensa del lanzamiento de la misión Gaia en el website de la Agencia Espacial Europea en url http://www.esa.int/esl/ESA_in_your_country/Spain/Dosier_de_Prensa_-_Mision_Gaia_-

(2) el término original en inglés (americano) es "billion", este fue traducido al español "mil millones", debido a que en español (según la Real Academia Española de la Lengua), "1 billón" equivale a "un millón de millones".

MAS NOTAS AL FINAL DEL DOCUMENTO.

lógica difusa para atacar el problema de escalabilidad en grandes volúmenes de datos [Ghosh & Mitra., 2013]. Otras alternativas propuestas en la literatura combinan los conceptos de lógica difusa y algoritmos genéticos para mejorar o solucionar en aplicaciones específicas los problemas de la eficiencia y escalabilidad.

En los llamados algoritmos inteligentes (algoritmos de aprendizaje), básicamente la principal razón para usar escalabilidad es que a medida que se incrementa el tamaño del conjunto de datos para el entrenamiento se incrementa a su vez la precisión de la clasificación aprendida de modelos. En la mayoría de los casos la pobre precisión cuando se usan algoritmos de aprendizaje es debido a muestras insuficientes (o que no contienen información relevante), presencia de ruido en los datos, o la existencia de un gran número de diferentes características. Así, usar escalabilidad en Minería de Datos implica que los algoritmos utilizados desarrollen un aprendizaje lo suficientemente rápido. Finalmente, el objetivo del aprendizaje (en cuanto a la precisión) no puede ser sacrificado por un algoritmo de escalamiento [Hu et al., 2003], es decir, existe un compromiso y debe existir un balance entre el escalamiento y el algoritmo usado.

El tema de la escalabilidad de la Minería de Datos en extensas bases de datos sigue siendo un campo de investigación de gran importancia. Como es bien sabido, el volumen de información día a día crece de manera exponencial, generando cada vez mayor cantidad de datos (con miles de atributos) almacenados, por lo que la necesidad de desarrollar algoritmos más eficientes y escalables está siempre entre las prioridades en las características de los algoritmos de búsqueda y [Riquelme et al. 2006, Kargupta et al. 2004].

## 4. SOLUCIONES DE MINERÍA DE DATOS EN ENTORNOS DISTRIBUIDOS

Los sistemas informáticos centralizados se difundieron en las décadas de los sesenta y setenta; luego con la aparición de los mini-ordenadores comenzaron a incorporarse procesos automatizados en diferentes campos de aplicación. Finalmente, la difusión masiva de los ordenadores personales en los ochenta y de las redes de comunicación generalizaron el uso de los procesos informáticos, obligando así a cambiar las estructuras centralizadas de los centros de proceso de datos [Caridad, 2001]. La mayoría de los datos comienzan a ser archivados en varias unidades de almacenamiento, es decir los datos se pueden almacenar en diferentes localidades lógicas, bien sea en un mismo espacio físico o geográficamente distinto. Estas bases de datos relacionadas lógicamente son interconectadas por una red de comunicaciones, dando lugar a las llamadas bases de datos distribuidas (BDD) [Mitra & Acharya, 2003].

En las BBD los múltiples computadores son llamados nodos (o sitios), y pueden tener diferentes esquemas de diseño, como por ejemplo el esquema centralizado, donde la BDD está localizada en una sola unidad de almacenamiento y los usuarios están distribuidos. El esquema de réplica, el cual consiste en que cada nodo debe tener su propia copia completa de la base de datos. Esquema fragmentado o particionado, donde solo hay una copia de cada elemento; en este caso la información está distribuida a través de los diferentes nodos y en cada nodo se almacena una o más partes disjuntas de la base de datos. Por último, podemos nombrar el esquema híbrido, el cual no es más que la combinación del esquema de partición y de réplica [Han & Kamber, 2006].

Las bases de datos distribuidas se pueden clasificar en homogéneas o heterogéneas. Las BDD homogéneas son aquellas en las que el mismo esquema de diseño está repetido en cada servidor y los registros se encuentran repartidos en los diferentes nodos. Mientras que, las BDD heterogéneas son aquellas en las que cada nodo almacena un subconjunto de datos [Mamani, 2015].

A su vez, el crecimiento de las bases de datos distribuidas dio origen a la necesidad de tratar con grandes cantidades de bases de datos heterogéneas, es decir datos almacenados en múltiples archivos, diferentes unidades de almacenamiento, y diferentes localidades geográficas, por lo que algoritmos más sofisticados en el uso de técnicas de Minería de Datos (MD) deben ser considerados para la integración de las BDD y la extracción de información de interés de las mismas [Mitra et al., 2003].

A la par con el desarrollo tecnológico, en el campo computacional, de sistemas con multiprocesadores, y el crecimiento de las de las BDD surge el llamado procesamiento paralelo. Los sistemas multiprocesadores pueden estar formados por sistemas distribuidos o sistemas centralizados de multiprocesadores, y pueden ser con memoria compartida o sin memoria compartida. Así este campo computacional de procesamiento paralelo ayuda en el proceso de velocidad de ejecución de los algoritmos y abre las puertas al desarrollo de nuevos algoritmos basados en esta herramienta de procesamiento paralelo. En un sistema de bases de datos paralelas, se pueden ejecutar varias operaciones y funciones a la par, por lo que estas técnicas de procesamiento paralelo se comienzan a considerar para mejorar el rendimiento en la Minería de Datos [García & Molina., 2012].

El rápido crecimiento de las bases de datos y comenzar estas a disponerse desde fuentes de almacenamiento distribuido donde los datos comúnmente son heterogéneos, semiestructurados o no estructurados, variantes en el tiempo y de alta dimensión (atributos) [Mitra et al., 2003] hacen que los métodos tradicionales de Minería de Datos, diseñados para bases de datos centralizadas, no sean suficientes para manejar las BDD; como por ejemplo, para manejar bases de datos en redes locales, redes inalámbricas, e Internet [Han & Kamber, 2006] , siendo ineludible el desarrollar algoritmos más avanzados que abarquen todas estas nuevas características, dando lugar al campo de la minería de datos distribuida (MDD).

La Minería de Datos Distribuida (MDD) se define como un mecanismo de análisis que consiste en la búsqueda y extracción de conocimiento oculto y no trivial en grandes volúmenes de bases de datos distribuidas entre varios nodos [Dubitzky, 2008] .La Minería de Datos Distribuida asume que los datos están distribuidos en dos o más nodos (sitios) y estos nodos cooperan entre sí para obtener resultados globales sin revelar los datos de cada nodo, o solo revelando partes de éstos [García et al., 2008]. En el proceso de MDD indicaciones pueden ser recibidas desde diferentes localidades,

---

(1) Se puede consultar el dosier de prensa del lanzamiento de la misión Gaia en el website de la Agencia Espacial Europea en url http://www.esa.int/esl/ESA_in_your_country/Spain/Dosier_de_Prensa_-_Mision_Gaia_-

(2) el término original en inglés (americano) es "billion", este fue traducido al español "mil millones", debido a que en español (según la Real Academia Española de la Lengua), "1 billón" equivale a "un millón de millones".

MAS NOTAS AL FINAL DEL DOCUMENTO.

al igual que estas pueden ser enviadas a varios destinos, por lo que los métodos de MDD deben ser usados para analizar las redes de conexión de datos provenientes de diferentes espacios geográficos.

Al igual que en el enfoque centralizado, se hacen uso de técnicas de Minería de Datos para la identificación de relaciones, patrones, asociaciones, segmentos, clasificaciones y tendencias, siendo en este caso sobre entornos distribuidos [Mamani, 2015]. Surgen nuevos algoritmos, como lo son los algoritmos paralelos y los algoritmos distribuidos en apoyo a la MDD. En general, estos algoritmos dividen los datos en particiones, las cuales son procesadas en paralelo, luego los resultados de estas particiones son integrados. Además, se desarrollan algoritmos incrementales, los cuales incorporan actualizaciones en las bases de datos sin tener la necesidad de realizar nuevas revisiones sobre toda la base de datos ya existente, lo cual reduce el costo computacional de las técnicas en MDD [Han & Kamber, 2006].

Existen tres estructuras clásicas de arquitectura para minería de datos distribuida:

La Estructura 1 consiste, como se muestra en la Figura 1, en que cada nodo distribuido dispone de un componente de minería encargado de minar los datos en la base de datos local, obteniéndose un modelo parcial para cada proceso de MD en cada uno de los nodos.

Todos los modelos parciales luego son integrados para obtener el modelo de Minería de Datos global [Mamani, 2015].

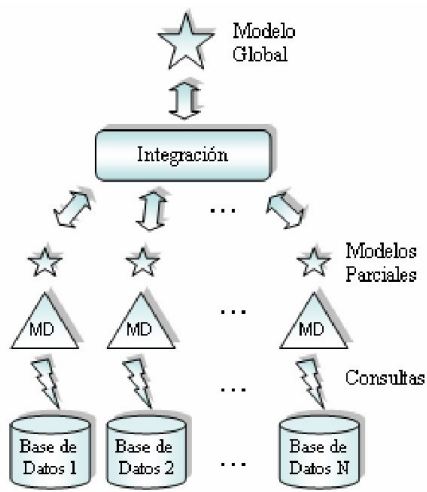

**Figura 2. Estructura 1 de la arquitectura clásica para MDD. Tomada de [Mamani, 2015].**

El principal beneficio de esta estructura en general se ve reflejado en el paralelismo que pueden aplicar sobre las bases de datos distribuidas.

El mismo algoritmo puede operar simultáneamente sobre cada nodo de datos distribuido, produciendo un modelo local por cada nodo. Por lo tanto, todos estos modelos parciales deben ser posteriormente integrados para producir el modelo final. En este caso de estructura, el éxito de los algoritmos de MDD se debe a la calidad de la integración de los modelos parciales.

Cada uno de estos modelos representa patrones relacionados localmente, pero pueden carecer de información necesaria para la obtención del modelo global [Park & Kargupta, 2003].

La Estructura 2 presentada en la Figura 3, consiste en realizar primeramente consultas en cada base de datos distribuida de manera independiente, las cuales son luego integradas. El proceso de Minería de Datos, en este caso, se realiza sobre la integración para formar el modelo global [Mamani, 2015].

Se puede observar que las Estructuras 2 y 3 (Figura 3 y Figura 4, respectivamente) realizan una sola vez el proceso de Minería de Datos, y este proceso se realiza sobre la integración de las bases de datos distribuidas. La diferencia entre estas dos estructuras se refleja en la forma en que se genera la integración sobre la que actúa la Minería de Datos [Mamani, 2015].

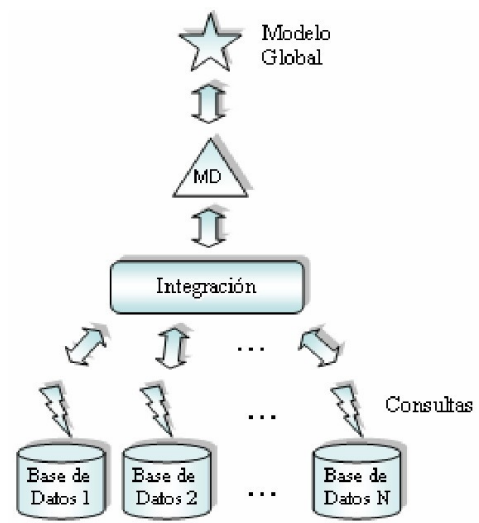

**Figura 3. Estructura 2 de la arquitectura clásica para MDD. Tomada de [Mamani, 2015].**

La Estructura 3, invierte el orden del proceso de consulta y de integración con respecto a la Estructura 2 ya presentada. Por lo que, primeramente, se integran todas las bases de datos distribuidas y las consultas se realizan a partir de la integración de los datos para el proceso de Minería de Datos, formando así un único modelo global, como se aprecia en la Figura 4 [Mamani, 2015].

Para compensar el problema de falta de información necesaria para la obtención del modelo global en la Estructura 1, es posible, como hacen algunos algoritmos de MDD, centralizar un subconjunto de los datos distribuidos. Sin embargo, minimizar la transferencia de datos es otra de las claves para el éxito de un algoritmo de MDD [Park & Kargupta, 2003]. Por lo que debe existir un balance entre el uso de los algoritmos y la estructura escogida.

La mayoría de las técnicas y algoritmos de minería de datos para BDD que se han propuesto, son modificaciones de técnicas clásicas de Minería de Datos para bases de datos relacionales, entre ella podemos nombrar [Mamani, 2015]:

---

(1) Se puede consultar el dosier de prensa del lanzamiento de la misión Gaia en el website de la Agencia Espacial Europea en
    url http://www.esa.int/esl/ESA_in_your_country/Spain/Dosier_de_Prensa_-_Mision_Gaia_-

(2) el término original en inglés (americano) es "billion", este fue traducido al español "mil millones", debido a que en español
    (según la Real Academia Española de la Lengua), "1 billón" equivale a "un millón de millones".

MAS NOTAS AL FINAL DEL DOCUMENTO.

En el caso de MDD en bases de datos homogéneas, donde por lo general se trabaja con la Estructura 1, los modelos parciales son tratados como si fueran modelos correspondientes a muestras diferentes, se usan técnicas como por ejemplo el meta-aprendizaje [6] para aprender meta-conceptos, basados en los conceptos localmente aprendidos. Por otro lado, son muy comunes el uso de las técnicas de aprendizaje bayesiano, donde considerando agentes bayesianos se estiman los parámetros de la distribución objetivo y una población de sistemas de aprendizaje que combinan las salidas de los modelos bayesianos producidos [Sierra, 2006].

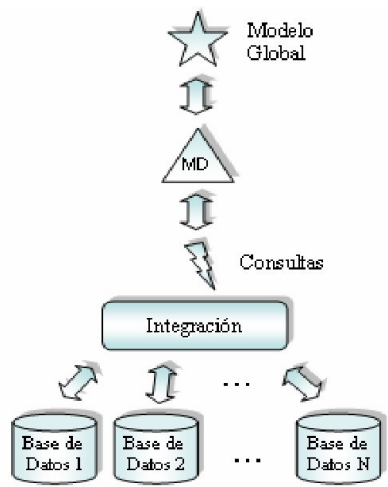

**Figura 4. Estructura 3 de la arquitectura clásica para MDD. Tomada de [Mamani, 2015].**

En el caso de la MDD heterogénea, tenemos la técnica llamada *Collective Data Modelling*, que es empleada para aprender árboles de decisión y para la regresión multivariante, consistente en la creación de modelos localmente correctos y la generación posterior del modelo global a través de la integración de los resultados locales, basada en la Estructura 1 [Hernández et al., 2004].

Existen otros métodos como multi-clasificadores, métodos de construcción de meta-modelos o métodos de combinación y ensamble de modelos que son basado en el principio: "si más de una hipótesis es consistente con los datos, se deben mantener todas (las hipótesis)" [Sierra, 2006].

También se han propuesto algoritmos de minería paralelos para trabajar con conjuntos de datos grandes, dividiéndolos y distribuyéndolos entre los distintos procesos de una máquina virtual. Uno de los métodos más intuitivos para encontrar reglas de asociación de manera distribuida se conoce como partición horizontal de datos, donde el proceso de minería se aplica localmente y los resultados obtenidos en cada nodo se combinan finalmente para obtener reglas que se cumplen en la mayoría de las bases de datos locales [García et al., 2008].

Una de las principales ventajas computacionales de los algoritmos genéticos (AG) es el permitir que sus operaciones se puedan ejecutar en paralelo, pudiendo distribuir las tareas en diferentes procesadores, por lo que resulta natural el aplicar este tipo de algoritmos a la MDD. Los llamados algoritmos genéticos paralelos (AGP) consisten en distribuir las tareas de un AG en diferentes procesadores. Al utilizar computadores en paralelo, no solamente se provee de más espacio de almacenamiento, sino también con el uso de ellos se pueden producir y evaluar más soluciones en menor tiempo. Debido al paralelismo, es posible incrementar el tamaño de la población, reducir el costo computacional y mejorar el desempeño de los AG. La idea básica es que los diferentes procesadores puedan crear nuevos individuos y computar sus aptitudes en paralelo, sin tener que comunicarse con los otros. La evaluación de la población en paralelo es simple de implementar, donde a cada procesador se le asigna un subconjunto de individuos para ser evaluados [Vélez & Silva, 2000].

A pesar de que los algoritmos paralelos son deseables para procesar de manera paralela las diferentes bases de datos, existe un compromiso y un balance que se debe tener en cuenta entre el esquema de comunicación, el uso de memoria, y la sincronización para que estos algoritmos sean eficientes [Kargupta et al., 2004], [Mitra et al., 2003]. Varias propuestas han surgido para solventar estos problemas, Xue presenta un algoritmo genético mejorado con reproducción asexual [Xue et al., 2012]. Para evitar redundancia binaria en los algoritmos, Kamimura propone un nuevo algoritmo genético distribuido que usa procesamiento paralelo [Kamimura & Kanasugi, 2012]. En (Hilda et al., 2015) se utilizan algoritmos genéticos paralelos para mejorar la efectividad en la selección de características y así reducir la dimensión (atributos) de las bases de datos.

Un sistema de MDD debe proporcionar un acceso eficiente a las diferentes fuentes de datos distribuidos, monitorear todo el proceso de minería de datos y mostrar los resultados al usuario en un formato apropiado, además de proveer una forma fácil de actualizar sus componentes. Por lo que avanzados esquemas de diseño han sido propuestos, como por ejemplo los modelos de arquitectura basados en la tecnología *grid* [7] y modelo de arquitectura basado en agentes [8] [Dubitzky, 2008].

Los modelos de arquitectura basados en la tecnología *grid* (celdas) soportan estructuras descentralizadas y paralelas para el análisis de bases de datos, ofreciendo así una alternativa para el manejo de datos en MDD, donde los algoritmos de MDD y el proceso del descubrimiento del conocimiento son complejos, computacionalmente hablando [Pérez et al., 2007]. Por otro lado, los sistemas basados en *grid* son bastante complejos, su naturaleza heterogénea y distribuida implica una complejidad arquitectónica de alto nivel. Las técnicas de MD contribuyen a observar y analizar los sistemas como un todo, brindando una capa de abstracción que reduce el nivel de observación en los sistemas basados en *grid*. Por lo que ambas técnicas se complementan, creando así un nuevo esquema de arquitectura para la MDD [Pérez et al., 2007]. A partir de la combinación de estas dos técnicas, han surgido una serie de algoritmos y metodologías para progresar y ofrecer alternativas ante las grandes demandas de aplicaciones en MDD y avances tecnológicos. [Ciubancan et al., 2013] expone algunas aplicaciones satisfactorias usando la combinación de MD con algoritmos basados en *grid*, por ejemplo, para en el campo de la ingeniería civil, la medicina y en telecomunicaciones. En [Slimani et al., 2014] se propone un nuevo algoritmo basado en la repetición de los datos en configuraciones *grid*, para mejorar el tiempo de respuesta, reducir el consumo del ancho de banda y mantener la

---

(1) Se puede consultar el dosier de prensa del lanzamiento de la misión Gaia en el website de la Agencia Espacial Europea en url http://www.esa.int/esl/ESA_in_your_country/Spain/Dosier_de_Prensa_-_Mision_Gaia_-

(2) el término original en inglés (americano) es "billion", este fue traducido al español "mil millones", debido a que en español (según la Real Academia Española de la Lengua), "1 billón" equivale a "un millón de millones".

MAS NOTAS AL FINAL DEL DOCUMENTO.

fiabilidad. Otro algoritmo, esta vez usando el comúnmente conocido algoritmo de agrupamiento (*clustering*) de datos, basado en *grid*, fue propuesto por [Tsai & Huang, 2015] para mejorar la eficiencia y efectividad en MD.

Recientemente ha cobrado interés el combinar el proceso de MDD con sistemas multiagentes [9]. En caso de los sistemas multiagente, estos tienen la necesidad de integrarse con la Minería de Datos debido al problema de mejorar la capacidad de aprendizaje de los agentes. Esto se logra mediante la alimentación de técnicas o algoritmos como: redes neuronales artificiales, razonamiento probabilístico, algoritmos genéticos, lógica difusa, arboles de decisión, entre otros. Incluir tecnología multiagente puede mejorar aspectos como: integración de varias fuentes de datos, acceso a aplicaciones distribuidas, interacción con diferentes usuarios y comunicación de diversas aplicaciones [Castillo & Meda, 2010]. Dentro de la MDD, son distintos los algoritmos, técnicas y aplicaciones en los que se pueden utilizar los sistemas multiagente, por ejemplo, [Zhou et al., 2010] presenta un nuevo algoritmo basado en agente para mejorar la inteligencia y eficiencia en la MDD basada en agentes. Khan [Khan & Mohammad, 2010] mejora el algoritmo de agrupamiento k-medias utilizando agentes para la generación de los centroides iniciales. Una técnica basada en detectar la norma de una comunidad de agentes locales, a través de un agente visitante es propuesta en [Mahmoud et al., 2012]. Qureshi [Qureshi et al., 2015] presenta una nueva arquitectura basada en multiagentes para la unificación de las reglas de asociación en MD. Una de las más recientes propuestas por Bu [Bu et al., 2016] expone un nuevo método basado en una perspectiva multiagente, este método utiliza los agentes para seleccionar la vecindad del nodo con la mayor similitud estructural que el nodo candidato, por lo que puede determinar si este debe ser añadido o no a la comunidad local.

Debido al gran campo de aplicación que hoy en día abarca la MDD, no es posible estar desarrollando algoritmos específicos acorde con un problema particular. La necesidad de ir a la generalización y de emplear estándares en Minería de Datos en inevitable. Este proceder simplifica la integración, actualización y mantenimiento de las aplicaciones y los sistemas que soportan tareas de Minería de [Mamani, 2015]. Estos estándares se ocupan de varios elementos, entre estos, la representación e intercambio de modelos estadísticos y de Minería de Datos, la representación y especificación de tareas de limpieza, transformación y agregación de los atributos usados como entradas en la creación de los modelos, la representación de parámetros internos requeridos para la construcción y uso de los modelos, las metodologías para el desarrollo de aplicaciones de minería de datos, las interfaces y las interfaces de programación de aplicaciones para garantizar el enlace necesario entre diferentes lenguajes y sistemas, y el análisis de datos remotos y distribuidos [García & Molina., 2012].

Existen enormes cantidades de bases de datos almacenadas de manera centralizada o distribuida, algunas áreas incluyen librerías digitales, bioinformática, biometría, finanzas, manufactura, producción, telecomunicaciones, ciencia [Mitra et al., 2003]. Sin embargo, dentro de la MDD, unas de las áreas con mayor demanda y crecimiento es la Web (WWW, *World Wide Web*), hoy en día considerada como la mayor base de datos distribuida multimedia, dando lugar al surgimiento de un campo más específico denominado *Web Mining* (minería de la Web), donde la MD utiliza técnicas avanzadas para recuperar información, extraer información, descifrar información sobre la estructura de relaciones (enlaces) y de los registro de navegación de los usuarios [Han & Kamber, 2006].

En conclusión, se puede decir que, la MDD es una disciplina de alto interés de los investigadores debido a las limitaciones que ofrecen las centralizadas. La mayoría de las técnicas y algoritmos de Minería de Datos para el análisis de fuentes de datos distribuidas homogéneas son extensiones de técnicas clásicas de Minería de Datos; sin embargo, el rápido crecimiento de bases de datos heterogéneas ha dado lugar al constante desarrollo de nuevas técnicas y algoritmos para MDD. Este campo seguirá creciendo a medida que crece el desarrollo tecnológico y las grandes bases de datos distribuidas.

## 6. CONCLUSIONES, IMPLICACIONES Y LIMITACIONES DEL ESTUDIO

Hasta hace relativamente poco las técnicas de minería de datos se utilizaban fundamentalmente en el tratamiento offline de grandes volúmenes de datos, bien para obtener información de valor a partir de los datos en bruto o para guiar algún proceso de soporte de decisiones. Con el incremento de la disponibilidad de los datos, y sobre todo de la mejora de las redes de telecomunicaciones ha sido posible la interconexión por primera vez de enormes bases de datos de contenido y naturaleza heterogénea lo que ha abierto nuevas perspectivas a la minería de datos que ahora es capaz de dar respuesta a sistemas de análisis de datos en tiempo real. La aparición de estándares y protocolos de interconexión de datos (XML, XSLT, JSON, HL7, DICOM), conjuntamente a la elaboración de estándares industriales (como PMML o CRISP-DM vistos más arriba) unido al desarrollo de la información *taggeada* y del contenido semántico (Ontologías, RDF, Web Semántica), al modelo de programación orientado a servicios (SOAP, REST, JAXRPC, WSDL, etc...) promete abrir a la minería de datos, conjuntamente con la madurez e implantación de las nuevas técnicas de aprendizaje automático, posibilidades de aplicación que hasta ahora no habían sido posible debido a su alto costo y complejidad computacional. Desde tele asistencia, navegación autónoma, enrutamiento automático, asistencia al diagnóstico, detección de intrusiones y rastreo de personas u objetos, predicción de indicadores en tiempo real, análisis de información sobre una red de computadores, interpretación del lenguaje natural, reconocimiento de escritura, vigilancia del tráfico aéreo, seguimiento de rastros en ordenadores e internet. Las posibilidades son muchas y en este sentido el procesamiento paralelo y el soporte a la computación distribuida se muestran clave para minimizar el impacto computacional y permitir la suficiente tolerancia del sistema para permitir su funcionamiento en un entorno cambiante como puede ser una red de ordenadores o internet.

**NOTAS A PIE DE PAGINA**

(1) Se puede consultar el dosier de prensa del lanzamiento de la misión Gaia en el website de la Agencia Espacial Europea en la url http://www.esa.int/esl/ ESA_in_your_country/ Spain/ Dosier_de_Prensa_-_Mision_Gaia_-

---

(1) Se puede consultar el dosier de prensa del lanzamiento de la misión Gaia en el website de la Agencia Espacial Europea en url http://www.esa.int/esl/ESA_in_your_country/Spain/Dosier_de_Prensa_-_Mision_Gaia_-

(2) el término original en inglés (americano) es "billion", este fue traducido al español "mil millones", debido a que en español (según la Real Academia Española de la Lengua), "1 billón" equivale a "un millón de millones".

MAS NOTAS AL FINAL DEL DOCUMENTO.

(2) el término original en inglés (americano) es "billion", este fue traducido al español "mil millones", debido a que en español (según la Real Academia Española de la Lengua), "1 billón" equivale a "un millón de millones".

(3) Algoritmo k-medias (*k-means algorithm*): es un método de agrupamiento, cuyo objetivo es segmentar de un conjunto de *n* observaciones k grupos, en el que cada observación pertenece al grupo más cercano a la media (Chadha & Kumar, 2014).

(4) Escalada Sigma: al comenzar la ejecución, los individuos más aptos no tendrán desviaciones estándar arriba de la media y por lo tanto no serán seleccionados. De igual manera, en la ejecución cuando la población es por lo general más convergente y la desviación estándar baja, los individuos más aptos permanecerán por fuera, permitiendo continuar la evolución (Vélez & Silva, 2000).

(5) Escalada por la Máxima Pendiente: En cada iteración se escoge aleatoriamente una solución, cercana a la solución actual, y si la seleccionada mejora la función de aptitud, esta se guarda. Este método converge a una solución óptima si la función de aptitud del problema es continua y si tiene solamente un pico (uni-modal). Si la función es multimodal (varios picos), el algoritmo termina en el primer pico que encuentre, aún sin ser éste el más alto (Vélez & Silva, 2000).

(6) Meta-aprendizaje: consiste en usar técnicas de aprendizaje supervisado para detectar conceptos en las bases de datos locales, para luego aprender meta-conceptos desde un conjunto de datos generados usando los conceptos localmente aprendidos (Sierra, 2006).

(7) El termino *grid*: en el campo de la computación, surge de la computación distribuida y los procesos de computo paralelos. Se trata de una infraestructura computacional distribuida que facilita la coordinación entre diferentes fuentes computacionales y/o geográficamente dispersas (Dubitzky, 2008).

(8) Los agentes: son entes lógicos o programas informáticos que perciben el entorno en el que están situados y a partir de tales percepciones ejecutan acciones de forma automática y flexible para dar solución colaborativa a problemas específicos previamente establecidos (Castillo & Meda, 2010).

(9) Sistema multi-agente: es la fusión/cooperación de varios agentes inteligentes que interactúan entre sí, con la finalidad de ejecutar tareas específicas de forma automática y flexible para dar solución a determinados problemas, en virtud de los objetivos establecidos (Castillo & Meda, 2010).

---

(1) Se puede consultar el dosier de prensa del lanzamiento de la misión Gaia en el website de la Agencia Espacial Europea en url http://www.esa.int/esl/ESA_in_your_country/Spain/Dosier_de_Prensa_-_Mision_Gaia_-

(2) el término original en inglés (americano) es "billion", este fue traducido al español "mil millones", debido a que en español (según la Real Academia Española de la Lengua), "1 billón" equivale a "un millón de millones".

MAS NOTAS AL FINAL DEL DOCUMENTO.

---

(1) Se puede consultar el dosier de prensa del lanzamiento de la misión Gaia en el website de la Agencia Espacial Europea en url http://www.esa.int/esl/ESA_in_your_country/Spain/Dosier_de_Prensa_-_Mision_Gaia_-

(2) el término original en inglés (americano) es "billion", este fue traducido al español "mil millones", debido a que en español (según la Real Academia Española de la Lengua), "1 billón" equivale a "un millón de millones".

MAS NOTAS AL FINAL DEL DOCUMENTO.

---

(1) Se puede consultar el dosier de prensa del lanzamiento de la misión Gaia en el website de la Agencia Espacial Europea en url http://www.esa.int/esl/ESA_in_your_country/Spain/Dosier_de_Prensa_-_Mision_Gaia_-

(2) el término original en inglés (americano) es "billion", este fue traducido al español "mil millones", debido a que en español (según la Real Academia Española de la Lengua), "1 billón" equivale a "un millón de millones".

MAS NOTAS AL FINAL DEL DOCUMENTO.

Conference on Information Engineering and Computer Science (ICIECS), pp. 1-4.

---

(1) Se puede consultar el dosier de prensa del lanzamiento de la misión Gaia en el website de la Agencia Espacial Europea en url http://www.esa.int/esl/ESA_in_your_country/Spain/Dosier_de_Prensa_-_Mision_Gaia_-

(2) el término original en inglés (americano) es "billion", este fue traducido al español "mil millones", debido a que en español (según la Real Academia Española de la Lengua), "1 billón" equivale a "un millón de millones".

MAS NOTAS AL FINAL DEL DOCUMENTO.